\documentclass{article}
\usepackage{array}
\usepackage{booktabs}
\usepackage{multirow}
\usepackage{rotating}
\usepackage{amsmath}
\usepackage{amssymb}
\usepackage{caption}
\usepackage{xcolor}
\usepackage{xspace}
\newcommand{\Ours}{EMem\xspace}
\newcommand{\OursGraph}{EMem-G\xspace}
\usepackage{tikz}
\usepackage{pgfplots}
\usetikzlibrary{patterns}
\pgfplotsset{compat=1.17}

\definecolor{babyblue}{rgb}{0.54, 0.81, 0.94}
\definecolor{blanchedalmond}{rgb}{1.0, 0.92, 0.8}
\definecolor{battleshipgrey}{rgb}{0.3, 0.3, 0.3}
\definecolor{brilliantrose}{rgb}{1.0, 0.33, 0.64}
\definecolor{americanrose}{rgb}{1.0, 0.01, 0.24}
\definecolor{jweigreen}{rgb}{0,0.45,0.24}
\definecolor{bluegray}{rgb}{0.1, 0.1, 0.4}
\definecolor{ao(english)}{rgb}{0.0, 0.5, 0.0}
\definecolor{blanchedalmond}{rgb}{1.0, 0.92, 0.8}
\definecolor{atomictangerine}{rgb}{1.0, 0.6, 0.4}
\definecolor{chocolate(web)}{rgb}{0.82, 0.41, 0.12}
\definecolor{bananayellow}{rgb}{1.0, 0.88, 0.21}
\definecolor{goldenbrown}{rgb}{0.6, 0.4, 0.08}
\definecolor{aliceblue}{rgb}{0.94, 0.97, 1.0}
\definecolor{beige}{rgb}{0.96, 0.96, 0.86}
\definecolor{babyblue}{rgb}{0.54, 0.81, 0.94}
\definecolor{camel}{rgb}{0.76, 0.6, 0.42}
\definecolor{cinnamon}{rgb}{0.82, 0.41, 0.12}
\usepackage{paralist}

\usepackage{microtype}
\usepackage{graphicx}
\usepackage{subfigure}
\usepackage{booktabs} 

\usepackage{hyperref}

\usepackage[accepted]{icml2025}

\usepackage{amsmath}
\usepackage{amssymb}
\usepackage{mathtools}
\usepackage{amsthm}

\usepackage[capitalize,noabbrev]{cleveref}

\theoremstyle{plain}

\theoremstyle{definition}

\theoremstyle{remark}

\usepackage[textsize=tiny]{todonotes}

\icmltitlerunning{A Simple Yet Strong Baseline for Long-Term Conversational Memory of LLM Agents}

\begin{document}

\twocolumn[
\icmltitle{A Simple Yet Strong Baseline for\\ Long-Term Conversational Memory of LLM Agents} 



\icmlsetsymbol{equal}{*}

\begin{icmlauthorlist}
\icmlauthor{Sizhe Zhou}{uiuc}
\icmlauthor{Jiawei Han}{uiuc}
\end{icmlauthorlist}

\icmlaffiliation{uiuc}{University of Illinois Urbana-Champaign}

\icmlcorrespondingauthor{Sizhe Zhou}{sizhez@illinois.edu}
\icmlcorrespondingauthor{Jiawei Han}{hanj@illinois.edu}

\icmlkeywords{Machine Learning, ICML}

\vskip 0.3in
]



\printAffiliationsAndNotice{}  

\begin{abstract}
LLM-based conversational agents still struggle to maintain coherent, personalized interaction over many sessions: 
fixed context windows limit how much history can be kept in view, and most external memory approaches trade off between coarse retrieval over large chunks and fine-grained but fragmented views of the dialogue. 
Motivated by neo-Davidsonian event semantics, we propose an event-centric alternative that represents conversational history as short, event-like propositions which bundle together participants, temporal cues, and minimal local context, rather than as independent relation triples or opaque summaries. 
In contrast to work that aggressively compresses or forgets past content, our design aims to preserve information in a non-compressive form and make it more accessible, rather than more lossy. 
Concretely, we instruct an LLM to decompose each session into enriched elementary discourse units (EDUs)---self-contained statements with normalized entities and source turn attributions---and organize sessions, EDUs, and their arguments in a heterogeneous graph that supports associative recall. 
On top of this representation we build two simple retrieval-based variants that use dense similarity search and LLM filtering, with an optional graph-based propagation step to connect and aggregate evidence across related EDUs. 
Experiments on the LoCoMo and LongMemEval$_\text{S}$ benchmarks show that these event-centric memories match or surpass strong baselines, while operating with much shorter QA contexts. 
Our results suggest that structurally simple, event-level memory provides a principled and practical foundation for long-horizon conversational agents. 
Our code and data will be released at \url{https://github.com/KevinSRR/EMem}.
\end{abstract}

\section{Introduction}\label{sec:introduction}

Large Language Models (LLMs) have demonstrated impressive conversational abilities, but their fixed context windows severely limit long-term coherence and personalization in extended interactions.  
Even in so–called long-context variants, performance can degrade sharply~\cite{liu-etal-2024-lost} and the LLMs can struggle to faithfully recall information that is many sessions old~\cite{locomo, wu2025longmemeval}. 
Meanwhile, naively concatenating entire multi-session histories into the prompt is computationally expensive and still bounded by a finite context window. 

A natural solution is to maintain an external store of dialogue history and retrieve relevant content. 
However, retrieval at the granularity of entire sessions or whole rounds often fails to recall fine-grained details while retrieval of turns fails to recover the larger context. 
Other approaches compress conversation histories into summaries or distilled “facts” before indexing~\cite{mem0, premem, licomemory, lightmem}, which improves efficiency but inevitably discards information. For long-term conversational memory---where queries may refer back to seemingly minor details many sessions ago---such lossy compression is risky.

To address these issues, recent research has drawn on cognitive science and structured knowledge representations. 
Cognitively-inspired architectures~\cite{nan2025nemori, amem, lightmem, li2025memos} organize memory into episodic and semantic stores, employ multi-stage consolidation, or treat memory as a managed system resource. 
In parallel, graph-based systems~\cite{hipporag, hipporag2, zep, sgmem, comorag, mem0} represent memories as networks of entities, relations, and text chunks, and perform graph search to support associative recall. 
These approaches highlight the importance of structure, yet most use entity--relation triples or coarse chunks as basic units, which fragments or coarsens the original discourse: a single utterance may be split into disconnected triples, while large chunks mix unrelated information.

Our starting point is the observation from neo-Davidsonian event semantics that the meaning of a sentence is often best represented as an \emph{event} with multiple arguments, rather than as a collection of independent binary relations~\cite{davidson_event-semantic, neo-Davidsonian_event-semantic}. 
Inspired by this view, we model long-term conversational memory at the level of \emph{event-like elementary discourse units (EDUs)}. 
Instead of taking raw clauses, we instruct an LLM to rewrite each session into a set of enriched EDUs. 
Unlike traditional EDUs from discourse parsing, our EDUs are short event-style statements that may span multiple utterances, enrich the text with normalized entities and minimal context information, and sometimes include lightly inferred information so that each EDU is as self-contained and precise as possible.\footnote{For example, from a dialogue about a trip to Tokyo we derive EDUs such as ``Bob traveled to Tokyo for five days to attend the Global AI Innovation Symposium 2024 in March 2024'' and ``Bob presented his team's multimodal learning work at the Global AI Innovation Symposium 2024.''. Here the entity mentions are normalized, and the timestamp is inferred based on the session timestamp. Note that the ``event'' concept also covers the expression of triple-style facts or knowledge. }
Together, these EDUs recover the information conveyed by the original session while making each atomic event explicit and maximally self-complete. 
We also attach turn-level source attributions to support downstream agentic behaviors to connect to larger and nuanced conversation contexts.

We then organize all sessions, EDUs, and extracted arguments from EDUs into an \emph{event-centric memory graph} to support associative recall and dense-sparse integration that are difficult to realize with flat retrieval over independent chunks. 
At query time, conversational questions often refer to unnamed or generic entities (\emph{“my pet”}, \emph{“that conference”}), so we perform entity and concept mention detection on the query and retrieve both EDUs and argument nodes in embedding space, using mentions as anchors into the graph rather than relying solely on exact entity strings. 
Because similarity scores and fixed thresholds alone are brittle in this setting, we employ lightweight LLM-based relevance filters over both EDUs and arguments, designed to favor recall, and use the resulting scores to define query-specific seed weights on the graph. 
The full model, \textbf{\OursGraph}, applies a Personalized PageRank step from these seeds to propagate relevance over EDU and argument nodes and select a small set of graph-consistent EDUs for augmenting the QA model, thereby capturing indirect associations across sessions.
The lightweight variant, \textbf{\Ours}, omits the graph and argument components and instead uses dense retrieval over EDUs followed by the same recall-oriented LLM filter, providing an efficient, conceptually simple baseline that still benefits from the event-centric representation.
In summary, our contributions are:
\begin{compactenum}
    \item We introduce an event-centric conversational memory representation based on EDUs and a heterogeneous graph linking sessions, EDUs, and argument nodes, grounded in neo-Davidsonian event semantics.
    \item We propose two retrieval variants: \OursGraph, which combines dense retrieval, LLM-based relevance filtering, and graph propagation via PPR; and \Ours, a lightweight dense-retrieval+filter baseline that avoids graph computation while retaining strong performance.
    \item We empirically demonstrate that these simple designs form competitive baselines on LoCoMo and LongMemEval$_\text{S}$.
\end{compactenum}

\section{Related Work}\label{sec:related_work}

\subsection{Memory Architectures for LLM Agents}

Cognitively-inspired memory systems such as Nemori~\cite{nan2025nemori}, LightMem~\cite{lightmem}, LiCoMemory~\cite{licomemory}, and A-Mem~\citep{amem} seek to emulate human memory processes by organizing interactions into episodes, applying multi-stage consolidation, and dynamically restructuring memories. 
Many of these frameworks perform explicit compression or abstraction---through clustering, summarization, or note-taking---to maintain a compact store, which improves efficiency but can drop fine-grained details. 
PREMem~\cite{premem} moves part of the reasoning to the write phase, storing enriched memory fragments that encode inferred relations across sessions. 
System-level frameworks like MemOS~\cite{li2025memos} and Mem0~\cite{mem0} treat memory as an operating system primitive or services layer, focusing on scalable storage and retrieval and often combining summarization with fact extraction. 
In contrast, our approach intentionally avoids lossy compression: we aim at memory representations that preserve all the original information and we rely on event-centric structure plus inference-time efforts to manage relevance and satisfy information seeking requests.

\begin{figure*}
    \centering
    \includegraphics[width=\linewidth]{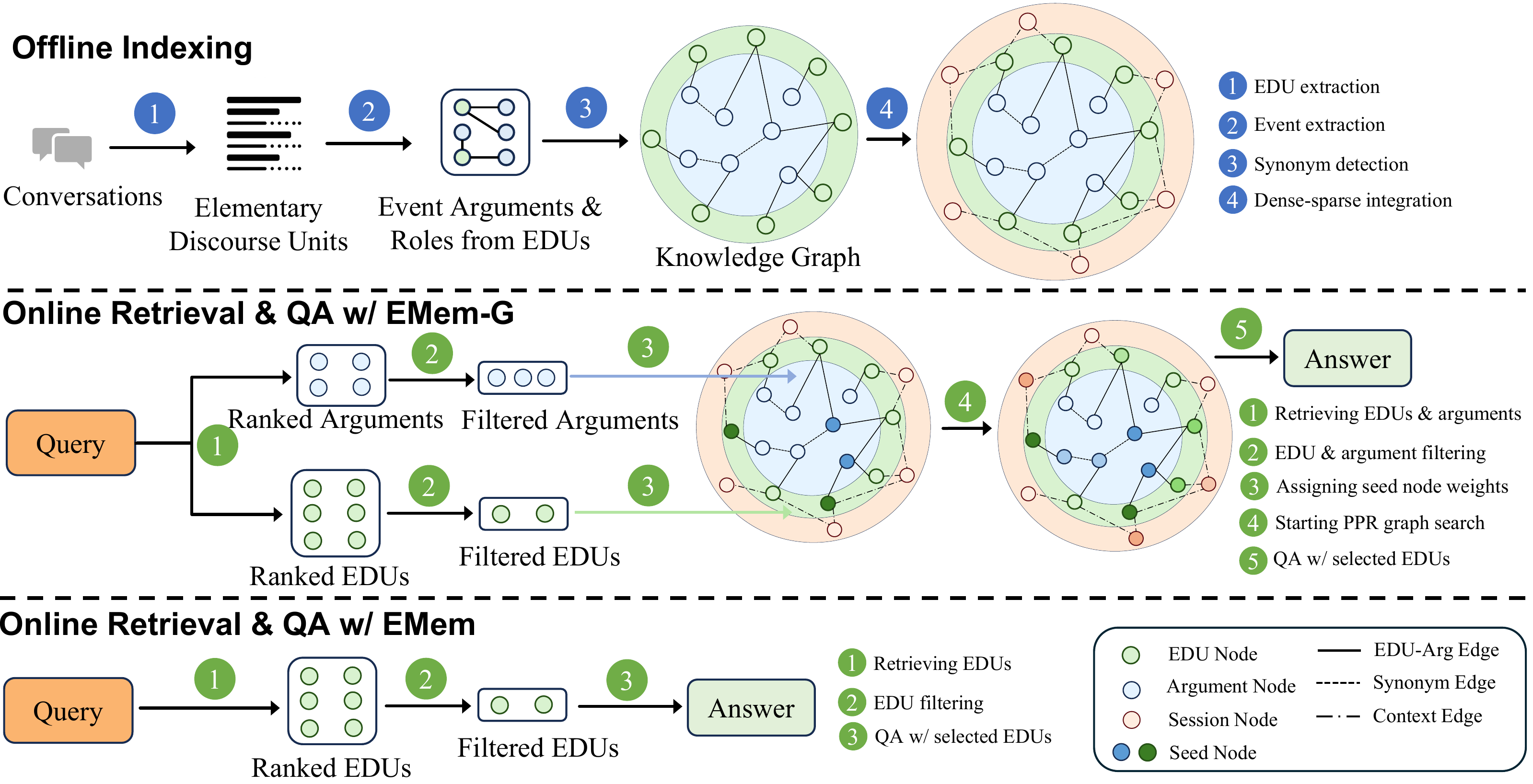}
      \vskip -0.1in
    \caption{Overview of our memory framework. Offline, conversations are decomposed into enriched EDUs, arguments, and an event graph; online, \OursGraph retrieves EDUs/arguments and propagates relevance over the graph, while \Ours uses only EDU retrieval and filtering to efficiently supply a compact memory context for QA.}
    \label{fig:methodology}
  \vskip -0.1in
\end{figure*}

\subsection{Structured and Graph-Based Memory}

Graph-based approaches have emerged as a powerful way to endow LLMs with associative recall. 
HippoRAG and HippoRAG 2~\cite{hipporag, hipporag2} build graphs over entities and passages and use personalized PageRank to retrieve multi-hop evidence. 
Zep~\cite{zep} and Mem0’s graph extension~\cite{mem0} maintain temporal knowledge graphs as a memory layer for agents, while ComoRAG~\cite{comorag} organizes long narratives into cognitively-inspired memory structures for iterative RAG. 
SGMem~\cite{sgmem} represents dialogue as sentence-level graphs, connecting sentences within and across sessions.
Our work is closest in spirit to HippoRAG 2 and SGMem but differs in two key aspects: 
(i) we adopt an \emph{event-centric} representation rooted in neo-Davidsonian semantics~\cite{davidson_event-semantic, neo-Davidsonian_event-semantic}, where enriched EDUs---rather than triples or raw sentences---are the primary memory units; 
and (ii) we couple graph-based retrieval with LLM-based recall-oriented filtering over both EDUs and argument nodes, which is particularly important for the implicit, unnamed references prevalent in less knowledge-intensive conversations.

\section{Methodology}\label{sec:methodology}

Figure~\ref{fig:methodology} illustrates the overall offline indexing and online retrieval pipelines.

\subsection{Problem Setting}

A conversation consists of sessions $\mathcal{S} = \{s_1,\dots,s_T\}$ ordered by timestamps $\tau(s)$. 
Each session $s$ is a sequence of turns
\(
s = \big((\textsc{spk}_{1}, u_{1}), \dots, (\textsc{spk}_{L_s}, u_{L_s})\big),
\)
where $u_\ell$ is the utterance from speaker $\textsc{spk}_{\ell}$. 
At query time, the agent receives a natural language question $q$ and answer it conditioned on the entire conversation history. 
We assume access to an embedding encoder $h(\cdot)$ that maps any text $x$ to an embedding $h(x)\in\mathbb{R}^d$, and a powerful QA model $f_{\text{QA}}$ that takes $(q,\text{memory})$ as input and generates an answer.

\subsection{Event-Centric Memory Graph}\label{sec:event_centric_memory_graph}

\paragraph{EDUs vs.\ relation triples.}
To make the notion of EDUs concrete, consider a short multi-turn exchange where the user describes a trip: from several utterances the extractor may produce an EDU such as \emph{``Bob spent five days in Tokyo in March~2024 to attend the Global AI Innovation Symposium~2024 at Tokyo University.''}
This EDU is mildly abstractive but self-contained: it unifies details mentioned across different turns and normalizes entities into their canonical forms.
Here, ``March~2024'' is inferred from the session timestamp and Bob’s utterance; and the mentions ``Global AI Innovation Symposium~2024'' and ``Tokyo University'' are expanded to full forms and normalized based on session context. 
As a memory item, this single EDU captures a complete event---who did what, where, when, and for what purpose.

In a triple-based knowledge graph, the same content would typically be decomposed into multiple relation triples, e.g.,
\((\textit{Bob}, \textit{attend}, \textit{Global AI Innovation Symposium 2024})\),
\((\textit{Bob}, \textit{stay\_in}, \textit{Tokyo})\),
\((\textit{Bob}, \textit{stay\_duration}, \textit{five days})\),
\((\textit{Symposium 2024}, \textit{held\_at}, \textit{Tokyo University})\),
\((\textit{Symposium 2024}, \textit{time}, \textit{March 2024})\),
which are stored as separate edges that may be scattered across the graph.  While such triples are useful for schema-driven reasoning, they fragment the original discourse: a retrieval step must find and recombine several triples to reconstruct the event, and fine-grained temporal or participant constraints can be lost or inconsistently represented.  In contrast, our enriched EDUs follow the neo-Davidsonian intuition of treating the event as a single unit with multiple arguments; each EDU node in the memory graph is therefore a self-complete, human-readable memory cell that preserves the local coherence of the original conversation while still being small enough to enable fine-grained retrieval.

\paragraph{EDU extraction.}
For each session $s$, we invoke an LLM-based extractor $g_{\text{EDU}}$ with a single in-context exemplar that describes the session (including the timestamp and all speaker names) and asks for a list of EDUs. 
The extractor outputs
\(
E_s = \{ e^{(s)}_1, \dots, e^{(s)}_{N_s} \},
\)
where each EDU $e$ is a short natural language description plus metadata:
\[
e = (\text{text}(e), \text{src}(e), \tau(e)).
\]
Here, $\text{src}(e)$ is the set of turn indices in $s$ that support $e$, and $\tau(e)$ is a timestamp derived from the session date (if available).
For long, structured assistant responses (e.g., enumerated suggestions) from LongMemEval$_\text{S}$, we allow the extractor to output \emph{structured chunks}: multi-sentence blocks, each accompanied by a 2--3 sentence summary that states the user request addressed, the information categories covered, and salient entities. 
We treat the summary as $\text{text}(e)$ for indexing and retrieval, and reserve the full chunk for the QA stage. 
For completeness, our dataset-specific treatment of long assistant responses in LongMemEval$_\text{S}$ is detailed in Appendix~\ref{appendix:longmemeval-edus}.
Across all sessions, the global EDU set is $\mathcal{E}=\bigcup_{s} E_s$.

\paragraph{Event-argument extraction.}
For each EDU $e\in\mathcal{E}$, we invoke a second LLM $g_{\text{ARG}}$ that treats $e$ as a single event and returns an event type $t(e)$ and a set of role--argument pairs
\(
\{(r_k, a_k)\}_{k=1}^{K_e}.
\)
We collect all unique argument strings into a global argument set $\mathcal{A}$. 
Each argument $a\in\mathcal{A}$ is associated with a node-level embedding $h_{\text{arg}}(a)$.
We do not enforce a fixed ontology of roles; the usage of $r_k$ is not explored in our framework, while arguments themselves become nodes in the memory graph.

\paragraph{Graph construction.}
We construct a heterogeneous graph
\(
G=(V,E)
\)
with three node types:
\begin{compactitem}
    \item Session nodes $v_s$ for $s\in\mathcal{S}$.
    \item EDU nodes $v_e$ for $e\in\mathcal{E}$.
    \item Argument nodes $v_a$ for $a\in\mathcal{A}$.
\end{compactitem}
Edges are defined as:
\begin{align*}
E_{\text{sess-edu}} &= \{(v_s, v_e) \mid e\in E_s\}, \\
E_{\text{edu-arg}} &= \{(v_e, v_a) \mid a \text{ is an argument of } e\}, \\
E_{\text{syn}} &= \{(v_a, v_{a'}) \mid \text{sim}(a,a') \ge \delta\}.
\end{align*}
Here, $\text{sim}(a,a')$ is cosine similarity between $h_{\text{arg}}(a)$ and $h_{\text{arg}}(a')$, and we cap the number of synonym neighbors per $a$ (e.g., at 100). 
The final node set is $V = \{v_s\}\cup\{v_e\}\cup\{v_a\}$ and edge set $E = E_{\text{sess-edu}}\cup E_{\text{edu-arg}}\cup E_{\text{syn}}$.

For later retrieval we cache embeddings for EDU texts, $h_{\text{edu}}(e)=h(\text{text}(e))$. 
Graph construction is performed offline as new sessions arrive.

\subsection{Graph-Based Retrieval and QA (\OursGraph)}\label{sec:method_oursgraph}

Given a query $q$, \OursGraph performs the following steps (corresponding to the middle row in Figure~\ref{fig:methodology}).

\paragraph{Dense retrieval of EDUs and arguments.}
We encode the query with the same encoder:
\(
z_q = h(q).
\)
We first retrieve the top-$K_e$ EDUs by cosine similarity between $z_q$ and $h_{\text{edu}}(e)$, obtaining a candidate set $C^{\text{edu}}(q)$. In parallel, we run an LLM-based mention detector on $q$ to extract a set of surface mentions $M(q)=\{m_1,\dots,m_M\}$ corresponding to entities, noun phrases, and salient concepts. Each mention is embedded as $h(m)$, and we retrieve top-$K_a$ argument nodes for each mention using similarity between $h(m)$ and $h_{\text{arg}}(a)$, forming a candidate argument set $C^{\text{arg}}(q)$.

\paragraph{Recall-oriented LLM filtering.}
Embedding similarity alone is brittle when mentions are generic (\emph{``pet''}, \emph{``that trip''}) or when arguments are highly specific. 
We therefore apply an LLM-based relevance filter to both candidate sets. For EDUs, we prompt a LLM once with the query $q$ and the list $\{\text{text}(e)\mid e\in C^{\text{edu}}(q)\}$, asking it to select the EDUs that are relevant to answering $q$. Its discrete selection induces a binary indicator
\(
f_{\text{EDU}}(q,e)\in\{0,1\},
\)
and we keep EDUs with $f_{\text{EDU}}(q,e)=1$. Similarly, for arguments we prompt the LLM with $q$ and the list $\{a\mid a\in C^{\text{arg}}(q)\}$ in minimal context, obtaining a binary indicator $f_{\text{ARG}}(q,a)\in\{0,1\}$. The filtered sets are
\begin{align*}
\tilde{C}^{\text{edu}}(q) &= \{ e\in C^{\text{edu}}(q)\mid f_{\text{EDU}}(q,e)=1\},\\
\tilde{C}^{\text{arg}}(q) &= \{ a\in C^{\text{arg}}(q)\mid f_{\text{ARG}}(q,a)=1\}.
\end{align*}
This design differs from precision-oriented filters (e.g., HippoRAG 2’s filter~\cite{hipporag2}): we intentionally keep borderline candidates by biasing the LLM toward recall, and rely on subsequent graph propagation and final QA to down-weight spurious ones.

\paragraph{Seed initialization.}
We initialize a nonnegative weight function $s:V\rightarrow\mathbb{R}_{\ge 0}$ over graph nodes using embedding similarities. For EDU nodes we set
\[
s(v_e) = \mathrm{sim}(z_q,h_{\text{edu}}(e)), \qquad v_e\in\tilde{C}^{\text{edu}}(q),
\]
and for argument nodes
\[
s(v_a) = \mathrm{sim}(h(m),h_{\text{arg}}(a)), \qquad v_a\in\tilde{C}^{\text{arg}}(q),
\]
with $s(v)=0$ for all remaining nodes and $m \in M(q)$ is the corresponding mention that retrieves this candidate argument $v_a$. 
If more than $K$ argument nodes receive nonzero scores, we keep only the $K$ highest-scoring ones and set the rest to zero, so that propagation remains focused and the initial mass is not diluted over many weakly related arguments. The resulting vector $\mathbf{s}$  is used as the personalization (seed) vector for personalized PageRank.

\paragraph{Personalized PageRank.}
Let $T$ be the column-stochastic transition matrix derived from $G$ (e.g., uniform over neighbors). We compute a Personalized PageRank vector
\[
\pi = \mathrm{PPR}(G,\mathbf{s}) \;\triangleq\; (1-\alpha)\mathbf{s} + \alpha T^\top \pi,
\]
with a fixed damping factor $\alpha\in(0,1)$ using a small number of power iterations. The resulting $\pi(v)$ scores reflect how strongly each node is connected to the query seeds under random walks that repeatedly return to $\mathbf{s}$.

\paragraph{Selecting EDUs and QA.}
We restrict $\pi$ to EDU nodes and select the top-$K$ EDUs,
\(
R(q) = \operatorname{TopK}\{(e,\pi(v_e)) : e\in\mathcal{E}\}.
\)
For EDUs corresponding to structured chunks, we replace $\text{text}(e)$ by the full chunk content before QA. We assemble a memory context by concatenating the selected EDUs along with their source session timestamps and source-turns' speaker names. 
Finally, the QA model produces the answer:
\[
\hat{y} = f_{\text{QA}}\big(q,\; \{(\text{text}(e),\text{src}(e),\tau(e)) : e\in R(q)\}\big),
\]
using a prompt that asks the model to first reason using the retrieved memories and then output a concise answer in zero-shot manner~\cite{cot}.

\subsection{Lightweight Retrieval (\Ours)}\label{sec:method_ours}

\Ours shares the same event graph and EDU extraction as \OursGraph but removes graph propagation and argument-level retrieval. 
Given a query $q$, we compute $z_q = h(q)$ and retrieve the top-$K_e$ EDUs by similarity to $h_{\text{edu}}(e)$, obtaining $C^{\text{edu}}(q)$. 
We then apply the same recall-oriented LLM filter $f_{\text{EDU}}$ as in Section~\ref{sec:method_oursgraph}: the LLM is prompted once with $q$ and the list $\{\text{text}(e) : e \in C^{\text{edu}}(q)\}$ and selects a subset of relevant EDUs, inducing binary decisions $f_{\text{EDU}}(q,e)\in\{0,1\}$. The retained set is
\[
R_{\text{lite}}(q) = \{ e\in C^{\text{edu}}(q)\mid f_{\text{EDU}}(q,e)=1\},
\]
so the number of EDUs passed to QA is determined adaptively by the filter rather than fixed a priori. The QA model then answers as
\[
\hat{y} = f_{\text{QA}}\bigl(q,\{(\text{text}(e),\text{src}(e),\tau(e)) : e\in R_{\text{lite}}(q)\}\bigr).
\]

Because EDUs are short, self-contained, and enriched with canonical entities and time information, dense retrieval over them already yields high-quality candidates. 
Retrieving a relatively large pool and then applying a recall-biased LLM filter captures most relevant memories without requiring personalized PageRank, while adaptively reducing the final context length. 
Empirically, this lightweight \Ours variant is competitive with \OursGraph and sometimes outperforms more sophisticated designs, making it a practical reference point for future work.

\section{Experiment}\label{sec:experiment}

\subsection{Experimental Setup}
\begin{table}[htb]
\centering
\small

\caption{Category distributions of the LoCoMo~\cite{locomo} and LongMemEval$_\text{S}$~\cite{wu2025longmemeval} datasets.}
\vspace{4pt}
\label{tab:dataset_stats}

\begin{tabular}{@{} l r  l r @{}} 

\toprule
\multicolumn{2}{c}{\textbf{LoCoMo}} & \multicolumn{2}{c}{\textbf{LongMemEval$_\text{S}$}} \\
\cmidrule(lr){1-2} \cmidrule(lr){3-4}
\textbf{Category} & \textbf{Count} & \textbf{Category} & \textbf{Count} \\
\midrule
Multi-Hop        & 278 & knowledge-update          & 72  \\
Temporal         & 320 & multi-session             & 121 \\
Open Domain      & 93  & single-session-assistant  & 56  \\
Single-Hop       & 829 & single-session-preference & 30  \\
--               & --  & single-session-user       & 64  \\
--               & --  & temporal-reasoning        & 127 \\
\midrule
\textbf{Total}   & \textbf{1,520} & \textbf{Total} & \textbf{470} \\
\bottomrule

\end{tabular}
\end{table}

\begin{table*}[t]
\caption{Performance on LoCoMo dataset~\cite{locomo} categorized by question type. Bold indicates the best performance. Underline indicates the second best performance. The baseline performance comes from \citealp{nan2025nemori}.}
\vspace{4pt}
\label{tab:main_results_locomo}
\resizebox{\linewidth}{!}{%
    \begin{tabular}{cl|ccc|ccc|ccc|ccc|ccc}
        \toprule
        & \multirow{2}{*}{\textbf{Method}} & \multicolumn{3}{c|}{\textbf{Temporal Reasoning}} & \multicolumn{3}{c|}{\textbf{Open Domain}} & \multicolumn{3}{c|}{\textbf{Multi-Hop}} & \multicolumn{3}{c|}{\textbf{Single-Hop}} & \multicolumn{3}{c}{\textbf{Overall}} \\
        \cmidrule{3-17}
        & & LLM Score & F1 & BLEU-1 & LLM Score & F1 & BLEU-1 & LLM Score & F1 & BLEU-1 & LLM Score & F1 & BLEU-1 & LLM Score & F1 & BLEU-1 \\
        \midrule
        \multirow{6}{*}{\rotatebox{90}{\textbf{gpt-4o-mini}}} & FullContext & $0.562 \pm 0.004$ & $0.441$ & $0.361$ & $0.486 \pm 0.005$ & $0.245$ & $0.172$ & $0.668 \pm 0.003$ & $0.354$ & $0.261$ & $0.830 \pm 0.001$ & $0.531$ & $0.447$ & $0.723 \pm 0.000$ & $0.462$ & $0.378$ \\
        \cmidrule(lr){2-17}
        & LangMem & $0.249 \pm 0.003$ & $0.319$ & $0.262$ & $0.476 \pm 0.005$ & $\mathbf{0.294}$ & $\underline{0.235}$ & $0.524 \pm 0.003$ & $0.335$ & $0.239$ & $0.614 \pm 0.002$ & $0.388$ & $0.331$ & $0.513 \pm 0.003$ & $0.358$ & $0.294$ \\
        & Mem0 & $0.504 \pm 0.001$ & $0.444$ & $0.376$ & $0.406 \pm 0.000$ & $0.271$ & $0.194$ & $0.603 \pm 0.000$ & $0.343$ & $0.252$ & $0.681 \pm 0.000$ & $0.444$ & $0.377$ & $0.613 \pm 0.000$ & $0.415$ & $0.342$ \\
        & RAG & $0.237 \pm 0.000$ & $0.195$ & $0.157$ & $0.326 \pm 0.005$ & $0.190$ & $0.135$ & $0.313 \pm 0.003$ & $0.186$ & $0.117$ & $0.320 \pm 0.001$ & $0.222$ & $0.186$ & $0.302 \pm 0.000$ & $0.208$ & $0.164$ \\
        & Zep & $0.589 \pm 0.003$ & $0.448$ & $0.381$ & $0.396 \pm 0.000$ & $0.229$ & $0.157$ & $0.505 \pm 0.007$ & $0.275$ & $0.193$ & $0.632 \pm 0.001$ & $0.397$ & $0.337$ & $0.585 \pm 0.001$ & $0.375$ & $0.309$ \\
        & Nemori & $0.710 \pm 0.000$ & $0.567$ & $0.466$ & $0.448 \pm 0.005$ & $0.208$ & $0.151$ & $0.653 \pm 0.002$ & $\underline{0.365}$ & $0.256$ & $0.821 \pm 0.002$ & $\mathbf{0.544}$ & $\mathbf{0.432}$ & $\underline{0.744} \pm 0.001$ & $\mathbf{0.495}$ & $0.385$ \\
        & \textbf{\OursGraph} & $\underline{0.760} \pm 0.003$ & $\mathbf{0.581}$ & $\mathbf{0.468}$ & $\underline{0.573} \pm 0.013$ & $0.242$ & $0.199$ & $\mathbf{0.747} \pm 0.006$ & $\mathbf{0.406}$ & $\underline{0.305}$ & $\underline{0.823} \pm 0.001$ & $\underline{0.504}$ & $\underline{0.422}$ & $\mathbf{0.780} \pm 0.000$ & $\underline{0.487}$ & $\mathbf{0.397}$\\
        & \textbf{\Ours} & $\mathbf{0.771} \pm 0.004$ & $\underline{0.574}$ & $\underline{0.461}$ & $\mathbf{0.602} \pm 0.009$ & $\underline{0.285}$ & $\mathbf{0.237}$ & $\underline{0.702} \pm 0.004$ & $\mathbf{0.406} $ & $\mathbf{0.307}$ & $\textbf{0.830} \pm 0.002$ & $0.497$ & $0.414$ & $\mathbf{0.780} \pm 0.001$ & $0.483$ & $\underline{0.393}$\\
        \midrule
        \multirow{6}{*}{\rotatebox{90}{\textbf{gpt-4.1-mini}}} & FullContext & $0.742 \pm 0.004$ & $0.475$ & $0.400$ & $0.566 \pm 0.010$ & $0.284$ & $0.222$ & $0.772 \pm 0.003$ & $0.442$ & $0.337$ & $0.869 \pm 0.002$ & $0.614$ & $0.534$ & $0.806 \pm 0.001$ & $0.533$ & $0.450$ \\
        \cmidrule(lr){2-17}
        & LangMem & $0.508 \pm 0.003$ & $0.485$ & $\underline{0.409}$ & $0.590 \pm 0.005$ & $\mathbf{0.328}$ & $\mathbf{0.264}$ & $0.710 \pm 0.002$ & $\underline{0.415}$ & $\textbf{0.325}$ & $0.845 \pm 0.001$ & $\underline{0.510}$ & $\underline{0.436}$ & $0.734 \pm 0.001$ & $\underline{0.476}$ & $\underline{0.400}$ \\
        & Mem0 & $0.569 \pm 0.001$ & $0.392$ & $0.332$ & $0.479 \pm 0.000$ & $0.237$ & $0.177$ & $0.682 \pm 0.003$ & $0.401$ & $0.303$ & $0.714 \pm 0.001$ & $0.486$ & $0.420$ & $0.663 \pm 0.000$ & $0.435$ & $0.365$ \\
        & RAG & $0.274 \pm 0.000$ & $0.223$ & $0.191$ & $0.288 \pm 0.005$ & $0.179$ & $0.139$ & $0.317 \pm 0.003$ & $0.201$ & $0.128$ & $0.359 \pm 0.002$ & $0.258$ & $0.220$ & $0.329 \pm 0.002$ & $0.235$ & $0.192$ \\
        & Zep & $0.602 \pm 0.001$ & $0.239$ & $0.200$ & $0.438 \pm 0.000$ & $0.242$ & $0.193$ & $0.537 \pm 0.003$ & $0.305$ & $0.204$ & $0.669 \pm 0.001$ & $0.455$ & $0.400$ & $0.616 \pm 0.000$ & $0.369$ & $0.309$ \\
        & Nemori & $0.776 \pm 0.003$ & $\mathbf{0.577}$ & $\mathbf{0.502}$ & $0.510 \pm 0.009$ & $0.258$ & $0.193$ & $0.751 \pm 0.002$ & $\mathbf{0.417}$ & $\underline{0.319}$ & $0.849 \pm 0.002$ & $\mathbf{0.588}$ & $\mathbf{0.515}$ & $0.794 \pm 0.001$ & $\mathbf{0.534}$ & $\mathbf{0.456}$ \\
        & \textbf{\OursGraph} & $\mathbf{0.808} \pm 0.001$ & $\underline{0.513}$ & $0.406$ & $\mathbf{0.717} \pm 0.005$ & $\underline{0.291}$ & $\underline{0.253}$ & $\mathbf{0.796} \pm 0.002$ & $0.376$ & $0.308$ & $\textbf{0.905} \pm 0.001$ & $0.510$ & $0.432$ & $\mathbf{0.853} \pm 0.000$ & $0.473$ & $0.393$\\
        & \textbf{\Ours} & $\underline{0.800} \pm 0.003$ & $0.491$ & $0.389$ & $\underline{0.652} \pm 0.010$ & $0.270$ & $0.237$ & $\underline{0.790} \pm 0.002$ & $0.350$ & $0.273$ & $\underline{0.897} \pm 0.001$ & $0.509$ & $0.430$ & $\underline{0.842} \pm 0.001$ & $0.461$ & $0.381$ \\
        \bottomrule
\end{tabular}%
}
\end{table*}
\begin{table*}
\centering
\caption{Performance on LongMemEval$_\text{S}$ dataset~\cite{wu2025longmemeval} across different question types. LLM-judged QA accuracy is reported. Bold indicates the best performance.}
\vspace{4pt}
\label{tab:main_results_longmemeval}
\footnotesize
\setlength{\tabcolsep}{3pt}
    \begin{tabular}{c|lcccc}
        \toprule
        & \multirow{2}{*}{\textbf{Question Type}} & \textbf{Full-context} & \textbf{Nemori} & \textbf{\OursGraph} & \textbf{\Ours} \\
        & & \textbf{(101K tokens)} & \textbf{(3.7-4.8K tokens)} & \textbf{(1.0K-3.6K tokens)} & \textbf{(0.6K-2.5K tokens)}\\
        \midrule
        \multirow{7}{*}{\rotatebox{90}{\textbf{gpt-4o-mini}}}
        & single-session-preference & 6.7\% & \textbf{46.7\%} & 32.2\% & 32.2\% \\
        & single-session-assistant & \textbf{89.3\%} & 83.9\% & 87.5\% & 82.1\% \\
        & temporal-reasoning & 42.1\% & 61.7\% & \textbf{74.8\%} & 69.8\% \\
        & multi-session & 38.3\% & 51.1\% & 73.6\% & \textbf{78.0\%} \\
        & knowledge-update & 78.2\% & 61.5\% & \textbf{94.4\%} & 87.5\% \\
        & single-session-user & 78.6\% & \textbf{88.6\%} & 87.0\% & 86.5\% \\
        & Average & 55.0\% & 64.2\% & \textbf{77.9\%} & 76.0 \% \\
        \midrule
        \multirow{7}{*}{\rotatebox{90}{\textbf{gpt-4.1-mini}}}
        & single-session-preference & 16.7\% & \textbf{86.7\%} & 50\% & 46.7\% \\
        & single-session-assistant & \textbf{98.2\%} & 92.9\% & 87.5\% & 82.1\% \\
        & temporal-reasoning & 60.2\% & 72.2\% & \textbf{83.7\%} & 80.6\% \\
        & multi-session & 51.1\% & 55.6\% & \textbf{82.6\%} & 82.1\% \\
        & knowledge-update & 76.9\% & 79.5\% & 94.4\% & \textbf{95.4\%} \\
        & single-session-user & 85.7\% & 90.0\% & \textbf{94.8\%} & 93.8\% \\
        & Average & 65.6\% & 74.6\% & \textbf{84.9\%} & 83.0\% \\
        \bottomrule
    \end{tabular}
\footnotesize
\end{table*}

We evaluated \Ours and \OursGraph on two widely used long-term conversational QA benchmark dataset, LoCoMo\footnote{The raw dataset provides four category labels (Categories~1--4), which we map to semantic categories: Category~1 to \textit{Multi-Hop}, Category~2 to \textit{Temporal Reasoning}, Category~3 to \textit{Open-Domain}, and Category~4 to \textit{Single-Hop} according to the discussion in \href{https://github.com/snap-research/locomo/issues/6}{this PR}.}~\cite{locomo} and LongMemEval$_\text{S}$~\cite{wu2025longmemeval}. 
LoCoMo contains 10 multi-session dialogues between two speakers with around 24K tokens on average while LongMemEval$_\text{S}$ consists of 500 multi-session dialogues between user and assistant with around 105K average tokens.
We exclude the adversarial type of questions following the prior research. 
The distribution of the evaluated questions are shown in Table~\ref{tab:dataset_stats}\footnote{Note that questions in LongMemEval$_\text{S}$ are assumed to be asked by the user. So we prepend the timestamp of the query and a prefix ``User: '' to each question in QA stage. An example processed query is ``Date of user query: 2023/05/30 (Tue) 21:54\textbackslash nUser: How old was I when my grandma gave me the silver necklace?''.}.

We follow Nemori~\cite{nan2025nemori} to set up the evaluation framework\footnote{Please refer to \href{https://github.com/nemori-ai/nemori/tree/main/evaluation}{the Nemori repository} for detailed implementation including the LLM-judge prompts.}, leveraging LLM-judge\footnote{The LLM-judge is based on \texttt{gpt-4o-mini}.} to score the final QA accuracy on two benchmark datasets, and additionally report the F1 and BLEU-1 scores for QA on the LoCoMo dataset. 
The LLM-judge runs three times with mean and standard deviation reported. 

We also follow Nemori~\cite{nan2025nemori} baseline setup to have: a full-context LLM that receives the entire dialogue history; a retrieval-augmented model (RAG-4096) that splits the dialogue into 4096-token chunks and uses dense retrieval to select context; and three memory-augmented systems, namely LangMem~\cite{langmem} with a hierarchical memory organization, Zep~\cite{zep} based on temporal knowledge graphs, and Mem0~\cite{mem0} which maintains a store of extracted personalized memories.

For \Ours and \OursGraph, we utilize the OpenAI \texttt{text-embedding-3-small} embedding model across all of our experiments, and use OpenAI \texttt{gpt-4o-mini} and \texttt{gpt-4.1-mini} as our backbone LLM respectively. 
In event graph construction, we set the similarity threshold $\delta$ for synonym edges to be $0.9$.
The number of initially retrieved EDUs top-$K_e$ is set to be $30$ and the number of initially retrieved arguments top-$K_a$ is set to $10$. 
The upper bound of the initialized argument nodes is $30$. 
The upper bound of the final retrieved EDUs top-$K$ is $10$.
For PPR, we followed HippoRAG 2~\cite{hipporag2}'s default parameters.

\subsection{Main Results}
\begin{table*}
\centering
\small
\caption{Statistics of graphs constructed by \OursGraph. All metrics except the first row are averaged per conversation. For LongMemEval$_\text{S}$, ``Avg Speaker EDU or Chunk Dist/Conv'' and ``Avg Speaker EDU or Chunk Len (words) Dist'' show the averaged count and averaged word length respectively in the format of ``User EDUs:Assistant EDUs:Assistant Chunks''. For LoCoMo, the two speakers of each conversation are ordered by their number of EDUs. The results are shown in the format of ``max-EDUs speaker:min-EDUs speaker''.}
\label{tab:graph_stats}
\vspace{4pt}
\resizebox{\linewidth}{!}{%
\begin{tabular}{lrrrr}
\toprule
 & \textbf{LongMemEval$_\text{S}$ (GPT-4o-mini)} & \textbf{LongMemEval$_\text{S}$ (GPT-4.1-mini)} & \textbf{LoCoMo (GPT-4o-mini)} & \textbf{LoCoMo (GPT-4.1-mini)} \\
\midrule
Number of Conversations & 470 & 470 & 10 & 10 \\
\midrule
Avg Sessions/Conv & 47.7 & 47.7 & 27.2 & 27.2 \\
Avg Session Length (words) & 1,644.4 & 1,644.4 & 536.9 & 536.9 \\
Avg Turns/Session & 10.3 & 10.3 & 21.6 & 21.6 \\
\midrule
Avg EDU Nodes/Conv & 861.1 & 1391.5 & 552.8 & 570.2 \\
Avg Arg Nodes/Conv & 2,780.9 & 3,786.2 & 1,144.7 & 1,184.9 \\
Avg Total Nodes/Conv & 3,689.7 & 5,225.4 & 1,724.7 & 1,782.3 \\
\midrule
Avg Session Node Degree & 18.1 & 29.2 & 20.3 & 21.0 \\
Avg EDU Node Degree & 5.5 & 4.7 & 4.9 & 4.7 \\
Avg Arg Node Degree & 1.5 & 1.4 & 1.9 & 1.8 \\
\midrule
Avg Session-EDU Edges/Conv & 861.3 & 1391.8 & 552.9 & 570.3 \\
Avg EDU-Arg Edges/Conv & 3,908.3 & 5,106.4 & 2,149.9 & 2,100.3 \\
Avg Synonym Edges/Conv & 105.8 & 189.0 & 24.5 & 38.7 \\
Avg Total Edges/Conv & 4,875.4 & 6,687.1 & 2,727.3 & 2,709.3 \\
\midrule
Avg User EDUs/Conv & 314.6 & 469.2 & -- & -- \\
Avg Asst EDUs/Conv & 421.6 & 725.6 & -- & -- \\
Avg Asst Chunks/Conv & 125.9 & 197.7 & -- & -- \\
Avg Speaker EDU or Chunk Dist/Conv & 314.6:421.6:125.9 & 469.2:725.6:197.7 & 98.1:16.7 & 105.9:18.5 \\
Avg Speaker EDU or Chunk Len (words) Dist & 23.2:21.3:193.8 & 20.3:21.1:121.3 & 15.0:15.4 & 16.9:17.9 \\
Avg Asst Chunk Summary Len (words) & 43.7 & 40.1 & -- & -- \\
\bottomrule
\end{tabular}
}
\end{table*}

Across both datasets and backbones, our method substantially improves over the memory baselines, while using comparable or fewer tokens (Tables~\ref{tab:main_results_locomo} and \ref{tab:main_results_longmemeval}). 
On LoCoMo, \Ours and \OursGraph consistently outperform in LLM-judged accuracy. 
With \texttt{gpt-4o-mini}, the overall LLM score improves from $0.744$ for Nemori to $0.780$ for both \Ours and \OursGraph. 
The gains concentrate on the categories that truly require long-term and structured reasoning: temporal reasoning, open-domain, multi-hop questions.
Single-hop questions are already easy for strong baselines, and all three top systems are effectively saturated there. 
The same pattern holds for \texttt{gpt-4.1-mini}: \OursGraph reaches $0.853$ overall vs.\ $0.806$ for full-context and $0.794$ for Nemori, indicating that memory retrieval can efficiently surpass naive full-context prompting even when the backbone LLM is strong. 
On LoCoMo, this accuracy is achieved with compact QA contexts: \Ours passes only $509$–$1039$ tokens to the backbone (average $738.2$), and \OursGraph $924$–$1062$ tokens (average $987.8$), which is substantially below Nemori’s reported $2{,}745$ tokens and far below the $23{,}653$-token full-context baseline~\cite{nan2025nemori}.
While Nemori remains slightly ahead in F1 on LoCoMo, \Ours/\OursGraph match or exceed Nemori on BLEU-1 and, more importantly, achieve higher LLM-judged correctness, suggesting that our answers are semantically more often correct even if word overlap is not always maximal.

The LongMemEval$_\text{S}$ results highlight the benefit of \Ours/\OursGraph in truly long conversations. 
With \texttt{gpt-4o-mini}, the average LLM-judged accuracy improves to 77.9\% for \OursGraph and 76.0\% for \Ours, while reducing the effective context from 101K tokens to 1.0K–3.6K and 0.6K–2.5K, respectively. 
With the stronger \texttt{gpt-4.1-mini}, we see the same trend. 
The largest gains come from temporal-reasoning, multi-session, and knowledge-update questions. 
These categories are exactly where the design of EDUs and event arguments matters: the EDU abstraction keeps who–did–what–when–where bundled into self-contained units, and recall-oriented LLM filtering plus event-aware retrieval makes it easier to locate and recombine the right long-range evidence than either full-context prompting or heuristic memory stores.

At the same time, Nemori remains competitive on single-session questions, especially preference and assistant-related ones.
These questions depend more on local stylistic cues and summarizing user habits within a short window than on integrating long-range event structure.
Nemori’s dual episodic–semantic memory, which first generates rich narrative episodes and then distills them into semantic knowledge (including user habits and inclinations), is therefore often sufficient and sometimes better aligned with the judge on these tasks.
By contrast, the EDU extractor is deliberately tuned toward factual, event-like content; as a result, purely attitudinal or stylistic information can be over-compressed or dropped, which limits performance on single-session-preference questions.

Comparing \Ours and \OursGraph directly, we see that the graph-based retrieval is helpful but not universally necessary. 
On LoCoMo, \Ours and \OursGraph attain essentially identical overall LLM scores with \texttt{gpt-4o-mini}, with \Ours slightly stronger on open-domain questions and \OursGraph slightly stronger on multi-hop questions. 
With \texttt{gpt-4.1-mini}, \OursGraph leads by about one percentage point overall. 
On LongMemEval$_\text{S}$, \OursGraph has a modest edge in average accuracy, and clearly helps on tasks that require stitching together scattered information (temporal reasoning, multi-session). 
In contrast, \Ours is often on par or slightly better on knowledge-update questions, where a small number of highly relevant, recent EDUs dominate and graph propagation adds limited additional signal.

Overall, these patterns suggest that (i) the event-semantics-centric EDU representation plus recall-oriented LLM filtering is the primary source of gains over baselines; (ii) graph-based propagation over arguments provides additional benefit mainly for queries that require relational and temporal integration across distant parts of the dialogue; and (iii) the lightweight \Ours variant is already a strong, practical default, with \OursGraph offering extra headroom on the most structurally demanding long-term memory tasks at a modest additional complexity cost.

\begin{table*}
\caption{Ablation performance on LoCoMo dataset~\cite{locomo} categorized by question type. \texttt{gpt-4o-mini} is adopted as base LLM and OpenAI \texttt{text-embedding-3-small} is adopted as base embedding model.}
\vspace{4pt}
\label{tab:ablation_locomo}
\resizebox{\linewidth}{!}{%
    \begin{tabular}{l|ccc|ccc|ccc|ccc|ccc}
        \toprule
        \multirow{2}{*}{} & \multicolumn{3}{c|}{\textbf{Temporal Reasoning}} & \multicolumn{3}{c|}{\textbf{Open Domain}} & \multicolumn{3}{c|}{\textbf{Multi-Hop}} & \multicolumn{3}{c|}{\textbf{Single-Hop}} & \multicolumn{3}{c}{\textbf{Overall}} \\
        \cmidrule{2-16}
        & LLM Score & F1 & BLEU-1 & LLM Score & F1 & BLEU-1 & LLM Score & F1 & BLEU-1 & LLM Score & F1 & BLEU-1 & LLM Score & F1 & BLEU-1 \\
        \midrule
        \textbf{\OursGraph} & $0.760 \pm 0.003$ & $0.581$ & $0.468$ & $0.573 \pm 0.013$ & $0.242$ & $0.199$ & $0.747 \pm 0.006$ & $0.406$ & $0.305$ & $0.823 \pm 0.001$ & $0.504$ & $0.422$ & $0.780 \pm 0.000$ & $0.487$ & $0.397$\\
        \ \ w/o Query MD to node & $0.754 \pm 0.006$ & $0.581$ & $0.468$ & $0.559 \pm 0.009$ & $0.276$ & $0.236$ & $0.675 \pm 0.007$ & $0.405$ & $0.305$ & $0.823 \pm 0.002$ & $0.507$ & $0.419$ & $0.766 \pm 0.001$ & $0.490$ & $0.397$\\
        \ \ \ \ \ w/ Query NED to node & $0.760 \pm 0.001$ & $0.580$ & $0.462$ & $0.559 \pm 0.009$ & $0.275$ & $0.226$ & $0.695 \pm 0.002$ & $0.407$ & $0.302$ & $0.820 \pm 0.000$ & $0.505$ & $0.420$ & $0.769 \pm 0.001$ & $0.489$ & $0.395$ \\
        \ \ w/o EDU Filter & $0.747 \pm 0.003$ & $0.565$ & $0.458$ & $0.516 \pm 0.000$ & $0.252$ & $0.201$ & $0.602 \pm 0.007$ & $0.357$ & $0.244$ & $0.796 \pm 0.001$ & $0.502$ & $0.420$ & $0.733 \pm 0.002$ & $0.473$ & $0.382$ \\ 
        \ \ w/o QA zero-shot CoT & $0.765 \pm 0.004$ & $0.595$ & $0.485$ & $0.595 \pm 0.005$ & $0.258$ & $0.208$ & $0.819 \pm 0.001$ & $0.413$ & $0.308$ & $0.819 \pm 0.001$ & $0.529$ & $0.443$ & $0.775 \pm 0.003$ & $0.505$ & $0.413$ \\ 
        \ \ w/o Graph \& PPR (\textbf{\Ours}) & $0.771 \pm 0.004$ & $0.574$ & $0.461$ & $0.602 \pm 0.009$ & $0.285$ & $0.237$ & $0.702 \pm 0.004$ & $0.406 $ & $0.307$ & $0.830 \pm 0.002$ & $0.497$ & $0.414$ & $0.780 \pm 0.001$ & $0.483$ & $0.393$ \\
        \ \ \ \ \ w/o EDU Filter & $0.748 \pm 0.005$ & $0.567$ & $0.462$ & $0.588 \pm 0.005$ & $0.268$ & $0.221$ & $0.633 \pm 0.010$ & $0.363$ & $0.252$ & $0.805 \pm 0.002$ & $0.502$ & $0.417$ & $0.748 \pm 0.003$ & $0.476$ & $0.384$ \\
        \ \ \ \ \ w/o QA zero-shot CoT & $0.768 \pm 0.004$ & $0.596$ & $0.476$ & $0.595 \pm 0.005$ & $0.291$ & $0.244$ & $0.701 \pm 0.003$ & $0.415$ & $0.307$ & $0.819 \pm 0.001$ & $0.520$ & $0.431$ & $0.773 \pm 0.001$ & $0.503$ & $0.407$ \\
        \bottomrule
\end{tabular}%
}
\end{table*}

\begin{table*}
\caption{Ablation performance on LongMemEval$_\text{S}$ dataset~\cite{wu2025longmemeval} across different question type. \texttt{gpt-4o-mini} is adopted as base LLM and OpenAI \texttt{text-embedding-3-small} is adopted as base embedding model. LLM-judged QA accuracy is reported.}
\vspace{4pt}
\label{tab:ablation_longmemeval}
\resizebox{\linewidth}{!}{%
    \begin{tabular}{l|c|c|c|c|c|c|c}
        \toprule
        & \multirow{2}{*}{\textbf{single-session-preference}} & \multirow{2}{*}{\textbf{single-session-assistant}} & \multirow{2}{*}{\textbf{temporal-reasoning}} & \multirow{2}{*}{\textbf{multi-session}} & \multirow{2}{*}{\textbf{knowledge-update}}  & \multirow{2}{*}{\textbf{single-session-user}} & \multirow{2}{*}{\textbf{Average}} \\
        & & & & & & \\
        \midrule
        \textbf{\OursGraph} & 32.2\% & 87.5\% & 74.8\% & 73.6\% & 94.4\% & 87.0\% & 77.9\% \\
        \ \ w/o Query MD to node & 26.7\% & 83.3\% & 72.7\% & 73.3\% & 88.9\% & 88.5\% & 75.8\% \\
        \ \ \ \ \ w/ Query NED to node & 26.7\% & 86.3\% & 76.1\% & 77.7\% & 91.7\% & 89.6\% & 78.8\% \\
        \ \ w/o QA zero-shot CoT & 3.3\% & 87.5\% & 72.4\% & 68.3\% & 90.3\% & 86.5\% & 73.4\% \\
        \ \ w/o EDU Filter & 10.0\% & 85.1\% & 74.5\% & 62.0\% & 90.3\% & 91.1\% & 73.1\% \\ 
        \ \ w/o Graph \& PPR (\textbf{\Ours}) &  32.2\% & 82.1\% & 69.8\% & 78.0\% & 87.5\% & 86.5\% & 76.0\% \\
        \ \ \ \ \ w/o EDU Filter & 26.7\% & 87.5\% & 70.1\% & 64.7\% & 84.7\% & 85.9\% & 72.4\% \\
         \ \ \ \ \ w/o QA zero-shot CoT & 15.5\% & 85.7\% & 56.7\% & 66.1\% & 94.4\% & 92.2\% & 70.6\% \\
        \bottomrule
\end{tabular}%
}
\end{table*}
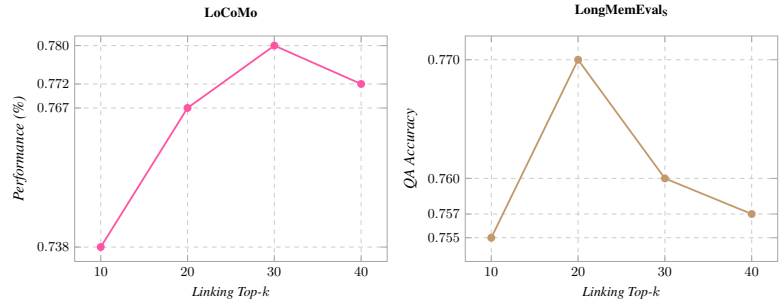
\begin{figure*}[t]
\centering
\resizebox{0.6\textwidth}{!}{%
\begin{tikzpicture}

\pgfplotsset{
    every axis/.style={
        width=0.42\textwidth,
        height=5.2cm,
        axis line style={very thin, gray!70},
        tick style={gray!70, thin},
        xlabel style={font=\small},
        ylabel style={font=\small},
        title style={font=\small, yshift=3pt},
        label style={font=\small},
        ticklabel style={font=\footnotesize},
        ymajorgrids=true,
        xmajorgrids=true,
        grid style={dashed, gray!40, line width=0.3pt},
        clip marker paths=true,
        scale only axis,
    }
}

\begin{axis}[
    name=leftplot,
    ymin=0.735, ymax=0.782,
    ytick={0.738,0.767,0.780,0.772},
    scaled y ticks=false,
    yticklabel style={
        /pgf/number format/.cd,
        fixed,
        precision=3,
        zerofill
    },
    xlabel={\emph{Linking Top-$k$}},
    ylabel={\emph{Performance (\%)}},
    ylabel near ticks,
    xtick={10,20,30,40},
    title={\textbf{LoCoMo}},
]

\addplot[
    color=brilliantrose,
    very thick,
    mark=*,
    mark size=2pt,
    mark options={fill=brilliantrose},
    line cap=round,
] coordinates {
    (10,0.738)
    (20,0.767)
    (30,0.780)
    (40,0.772)
};

\end{axis}

\begin{axis}[
    name=rightplot,
    at={(leftplot.east)},
    anchor=west,
    xshift=1.8cm,
    ymin=0.753, ymax=0.772,
    ytick={0.755,0.770,0.760,0.757},
    scaled y ticks=false,
    yticklabel style={
        /pgf/number format/.cd,
        fixed,
        precision=3,
        zerofill
    },
    xlabel={\emph{Linking Top-$k$}},
    ylabel={\emph{QA Accuracy}},
    ylabel near ticks,
    xtick={10,20,30,40},
    title={\textbf{LongMemEval$_\text{S}$}},
]

\addplot[
    color=camel,
    very thick,
    mark=*,
    mark size=2pt,
    mark options={fill=camel},
    line cap=round,
] coordinates {
    (10,0.755)
    (20,0.770)
    (30,0.760)
    (40,0.757)
};

\end{axis}

\end{tikzpicture}
}
\caption{LLM-judged QA accuracy of \Ours with different linking Top-$k$ setup. \texttt{gpt-4o-mini} is adopted as base LLM and OpenAI \texttt{text-embedding-3-small} is adopted as base embedding model.}
\label{fig:ours_qa_acc_w_varying_link_topk}
\end{figure*}
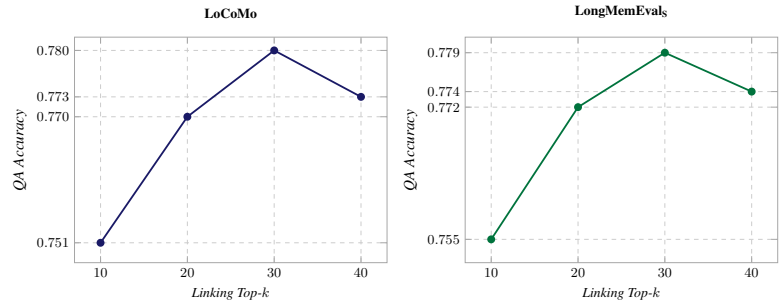
\begin{figure*}
\centering
\resizebox{0.6\textwidth}{!}{%
\begin{tikzpicture}

\pgfplotsset{
    every axis/.style={
        width=0.42\textwidth,
        height=5.2cm,
        axis line style={very thin, gray!70},
        tick style={gray!70, thin},
        xlabel style={font=\small},
        ylabel style={font=\small},
        title style={font=\small, yshift=3pt},
        label style={font=\small},
        ticklabel style={font=\footnotesize},
        ymajorgrids=true,
        xmajorgrids=true,
        grid style={dashed, gray!40, line width=0.3pt},
        clip marker paths=true,
        scale only axis,
    }
}

\begin{axis}[
    name=leftplot,
    ymin=0.748, ymax=0.782,
    ytick={0.751,0.770,0.780,0.773},
    scaled y ticks=false,
    yticklabel style={
        /pgf/number format/.cd,
        fixed,
        precision=3,
        zerofill
    },
    xlabel={\emph{Linking Top-$k$}},
    ylabel={\emph{QA Accuracy}},
    ylabel near ticks,
    xtick={10,20,30,40},
    title={\textbf{LoCoMo}}
]

\addplot[
    color=bluegray,
    very thick,
    mark=*,
    mark size=2pt,
    mark options={fill=bluegray},
    line cap=round,
] coordinates {
    (10,0.751)
    (20,0.770)
    (30,0.780)
    (40,0.773)
};

\end{axis}

\begin{axis}[
    name=rightplot,
    at={(leftplot.east)},
    anchor=west,
    xshift=1.8cm,
    ymin=0.752, ymax=0.781,
    ytick={0.755,0.772,0.779,0.774},
    scaled y ticks=false,
    yticklabel style={
        /pgf/number format/.cd,
        fixed,
        precision=3,
        zerofill
    },
    xlabel={\emph{Linking Top-$k$}},
    ylabel={\emph{QA Accuracy}},
    ylabel near ticks,
    xtick={10,20,30,40},
    title={\textbf{LongMemEval$_\text{S}$}},
]

\addplot[
    color=jweigreen,
    very thick,
    mark=*,
    mark size=2pt,
    mark options={fill=jweigreen},
    line cap=round,
] coordinates {
    (10,0.755)
    (20,0.772)
    (30,0.779)
    (40,0.774)
};

\end{axis}

\end{tikzpicture}
}
\caption{LLM-judged QA accuracy of \OursGraph with different linking Top-$k$ setup. \texttt{gpt-4o-mini} is adopted as base LLM and OpenAI \texttt{text-embedding-3-small} is adopted as base embedding model.}
\label{fig:ours_graph_qa_acc_w_varying_link_topk}
\end{figure*}
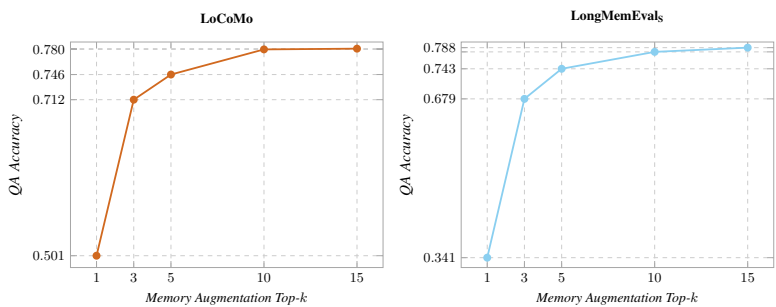
\begin{figure*}
\centering
\resizebox{0.6\textwidth}{!}{%
\begin{tikzpicture}

\pgfplotsset{
    every axis/.style={
        width=0.42\textwidth,
        height=5.2cm,
        axis line style={very thin, gray!70},
        tick style={gray!70, thin},
        xlabel style={font=\small},
        ylabel style={font=\small},
        title style={font=\small, yshift=3pt},
        label style={font=\small},
        ticklabel style={font=\footnotesize},
        ymajorgrids=true,
        xmajorgrids=true,
        grid style={dashed, gray!40, line width=0.3pt},
        clip marker paths=true,
        scale only axis,
    }
}

\begin{axis}[
    name=leftplot,
    ymin=0.485, ymax=0.790,
    ytick={0.501,0.712,0.746,0.780},
    scaled y ticks=false,
    yticklabel style={
        /pgf/number format/.cd,
        fixed,
        precision=3,
        zerofill
    },
    xlabel={\emph{Memory Augmentation Top-$k$}},
    ylabel={\emph{QA Accuracy}},
    ylabel near ticks,
    xtick={1,3,5,10,15},
    title={\textbf{LoCoMo}},
    extra y ticks={0.781},
    extra y tick labels={\empty}
]

\addplot[
    color=chocolate(web),
    very thick,
    mark=*,
    mark size=2pt,
    mark options={fill=chocolate(web)},
    line cap=round,
] coordinates {
    (1,0.501)
    (3,0.712)
    (5,0.746)
    (10,0.780)
    (15,0.781)
};

\end{axis}

\begin{axis}[
    name=rightplot,
    at={(leftplot.east)},
    anchor=west,
    xshift=1.8cm,
    ymin=0.320, ymax=0.800,
    ytick={0.341,0.679,0.743,0.788},
    scaled y ticks=false,
    yticklabel style={
        /pgf/number format/.cd,
        fixed,
        precision=3,
        zerofill
    },
    xlabel={\emph{Memory Augmentation Top-$k$}},
    ylabel={\emph{QA Accuracy}},
    ylabel near ticks,
    xtick={1,3,5,10,15},
    title={\textbf{LongMemEval$_\text{S}$}},
    extra y ticks={0.779},
    extra y tick labels={\empty}
]

\addplot[
    color=babyblue,
    very thick,
    mark=*,
    mark size=2pt,
    mark options={fill=babyblue},
    line cap=round,
] coordinates {
    (1,0.341)
    (3,0.679)
    (5,0.743)
    (10,0.779)
    (15,0.788)
};

\end{axis}

\end{tikzpicture}
}
\caption{LLM-judged QA accuracy of \Ours with different memory augmentation Top-$k$ setup. \texttt{gpt-4o-mini} is adopted as base LLM and OpenAI \texttt{text-embedding-3-small} is adopted as base embedding model.}
\label{fig:ours_graph_qa_acc_w_varying_qa_topk}
\end{figure*}
\subsection{Graph Statistics}\label{sec:graph_statistics}

Table~\ref{tab:graph_stats} summarizes the graphs constructed by \OursGraph and confirms that both benchmarks are structurally challenging. Conversations span dozens of sessions with long sessions in terms of word count, especially on LongMemEval$_\text{S}$, where assistant turns are very long. After EDU abstraction, however, each conversation is reduced to a manageable graph with on the order of $500$–$1{,}400$ EDUs and $1{,}000$–$4{,}000$ argument nodes, depending on the dataset and backbone. 
Moving from \texttt{gpt-4o-mini} to \texttt{gpt-4.1-mini} produces a large increase in EDUs and arguments on LongMemEval$_\text{S}$ but only a modest change on LoCoMo, indicating that stronger extractors matter most when turns are long and heterogeneous.

Despite the scale, the graphs are sparse: EDU nodes connect to only a handful of arguments, argument nodes have degree around one or two, and synonym edges are comparatively rare. 
This means that events form small, well-localized neighborhoods, which is favorable for personalized PageRank—random walks remain concentrated around a few relevant clusters rather than diffusing through a dense graph. 
This could also indicate that a better event argument extraction method producing more normalized, atomic arguments could form a more dense graph for better graph-based retrieval performance. 
The speaker-wise statistics further show that the number of memory units per role is both manageable and highly imbalanced: in LongMemEval$_\text{S}$ the assistant contributes the majority of EDUs and chunks, and in LoCoMo one interlocutor dominates the EDU count, reflecting where most factual content actually resides. 
At the same time, individual EDUs are short (roughly one to two sentences) and long assistant responses are represented by compact summaries, so most of the information that retrieval and QA operate on is concentrated into concise, event-centric nodes rather than raw turns. 
These properties together explain why our event-centric graph scales to very long conversations while still enabling efficient, focused retrieval.

\subsection{Ablation Study}\label{sec:ablation_study}

We ablate the main components of \Ours and \OursGraph on LoCoMo and LongMemEval$_\text{S}$ (Tables~\ref{tab:ablation_locomo} and~\ref{tab:ablation_longmemeval}). 
On LoCoMo, the LLM-based EDU filter is the most critical piece: removing it reduces the overall LLM score from $0.780$ to $0.733$ for \OursGraph and to $0.748$ for \Ours, with multi-hop performance dropping by $7$–$15$ points, confirming that recall-oriented filtering is key to suppressing noisy EDUs while preserving relevant events. 
Removing query mention--argument node weight initialization in \OursGraph leads to a smaller overall drop ($0.780 \rightarrow 0.766$) but hurts multi-hop reasoning ($0.747 \rightarrow 0.675$); 
replacing the LLM-based mention detector with a named entity mention variant largely closes this gap, indicating that graph propagation benefits from some form of query–argument anchoring but is fairly robust to the choice of detector. 
Dropping QA chain-of-thought has only a minor effect on the LoCoMo average (less than one point) for both models, mainly trading a small decrease in LLM score for slightly higher F1/BLEU.

On LongMemEval$_\text{S}$, the same components remain important but their impact is larger. 
For \OursGraph, removing the EDU filter or QA CoT reduces the average accuracy from $77.9\%$ to $73.1\%$ and $73.4\%$, respectively, with multi-session questions dropping by up to $11.6$ points, showing that pruning and explicit reasoning both matter when information is spread over many sessions. 
This could be related to our EDU extraction which leads to lots of topic-wise similar EDUs from same sessions, hence requiring a relatively more powerful embedding model, reranker, or filter to remove noisy information. 
Removing the graph and PPR (\Ours) lowers the average to $76.0\%$: the graph particularly helps temporal-reasoning and knowledge-update questions, while \Ours slightly improves multi-session performance, consistent with the trade-offs in the main results. 
For \Ours, ablations of the EDU filter and QA CoT again mainly harm multi-session and temporal-reasoning questions. 
Overall, these results support our design choice that event-semantics based EDUs plus recall-oriented filtering carry most of the gains, with graph propagation and QA CoT providing additional, dataset-dependent improvements on the hardest long-term reasoning cases.

\subsection{Retrieval Hyperparameters Anaysis}\label{sec:retrieval_hyperparam_analysis}

We first vary the number of retrieved candidate EDUs before filtering (linking Top-$K_e$) in \Ours (Figure~\ref{fig:ours_qa_acc_w_varying_link_topk}). Across both LoCoMo and LongMemEval$_\text{S}$, performance is relatively flat once $K_e$ is in the range of 20–40, with a mild optimum around 20–30 depending on the dataset. 
This indicates that the combination of dense retrieval and recall-oriented EDU filtering is robust: as long as we retrieve a moderately sized candidate pool, the LLM filter is able to discard noisy EDUs and preserve most relevant information, and there is no need to aggressively tune $K_e$.

For \OursGraph, we similarly vary linking Top-$K_e$, which also determines the upper bound on non-zero argument seeds (see Figure~\ref{fig:ours_graph_qa_acc_w_varying_link_topk}). 
Accuracy improves as $K_e$ grows from small values and peaks around 30 on both datasets, with only marginal degradation beyond that point, suggesting that graph propagation benefits from a richer set of seeds but becomes slightly more susceptible to noise when too many low-relevance EDUs and arguments are injected. 
Finally, varying the QA Top-$K$ (the number of EDUs passed to the QA model) for \OursGraph with fixed linking $K_e=30$ (Figure~\ref{fig:ours_graph_qa_acc_w_varying_qa_topk}) shows a steep gain when moving from very small contexts (1–3 EDUs) to moderate ones (5–10 EDUs), after which performance quickly saturates around 10–15 EDUs. 
Overall, these trends indicate that our memory system is stable across a broad range of retrieval and augmentation hyperparameters, and that strong performance can be obtained with small, fixed context budgets that remain practical for real-world deployment.

\section{Conclusion}

We have argued for an event-centric view of conversational memory, grounded in neo-Davidsonian semantics, where long-term dialogue is reconstructed as a graph of enriched EDUs rather than as raw turns, coarse summaries, or fragmented triples. 
By instructing an LLM to produce self-contained, normalized event units and organizing them into a heterogeneous graph over sessions, EDUs, and arguments, our framework enables associative recall and dense–sparse integration that are difficult to realize with flat retrieval over text chunks. 
The two retrieval variants, \Ours and \OursGraph, differ only in whether they invoke graph propagation, but share the same design philosophy: use simple, high-recall dense retrieval, apply a lightweight LLM filter to remove noisy candidates, and then reason over a compact set of event memories.

Experiments on LoCoMo and LongMemEval$_\text{S}$ show that this design yields strong performance on temporal, multi-hop, and knowledge-update questions while operating with modest context budgets, and that the induced memory graphs are sparse, interpretable, and robust to retrieval hyperparameters. 
At the same time, weaker performance on single-session preference questions highlights a limitation of a purely event-centric representation for capturing fine-grained user attitudes and styles. 
An important direction for future work is to pair event-level memory with complementary models of user profiles and interaction patterns, and to extend this framework beyond dialogue to other long-horizon settings such as tool-augmented agents and multi-document reasoning.

\section*{Acknowledgments}

Research was supported in part by the AI Institute for Molecular Discovery, Synthetic Strategy, and Manufacturing: Molecule Maker Lab Institute (MMLI), funded by U.S.\ National Science Foundation under Awards No.\ 2019897 and 2505932, NSF IIS 25-37827, and the Institute for Geospatial Understanding through an Integrative Discovery Environment (I-GUIDE) by NSF under Award No.\ 2118329.  The research has used the Delta/DeltaAI advanced computing and data resource, supported  in part by the University of Illinois Urbana-Champaign and through allocation \#250851 from the Advanced Cyberinfrastructure Coordination Ecosystem: Services \& Support (ACCESS) program, which is supported by National Science Foundation grants OAC 2320345, \#2138259, \#2138286, \#2138307, \#2137603, and \#2138296.  Any opinions, findings, and conclusions or recommendations expressed herein are those of the authors and do not necessarily represent the views, either expressed or implied, of DARPA or the U.S.\ Government.
The views and conclusions contained in this paper are those of the authors and should not be interpreted as representing any funding agencies.

\bibliography{example_paper}
\bibliographystyle{icml2025}

\newpage
\appendix
\onecolumn

\section{Dataset-Specific EDU Extraction for LongMemEval}\label{appendix:longmemeval-edus}

LongMemEval consists of multi-session dialogues between a \emph{user} and an \emph{assistant}~\citep{wu2025longmemeval}. 
We observed a systematic asymmetry in utterance style: user turns are typically short and focused, whereas assistant turns are often long, highly structured responses (e.g., enumerated lists, step-by-step plans, comparative analyses). 
Leveraging a relatively small LLMs (e.g., \texttt{gpt-4o-mini}) for EDU extraction on such long sessions, we frequently encounter the issue of missing EDUs, which is likely considered as unimportant information. 
To adapt our EDU extraction pipeline to this setting without overfitting the core method, we apply a slightly different treatment to the two speakers.

\paragraph{User utterances.}
For the user side, we directly apply the generic EDU extraction procedure from Section~\ref{sec:event_centric_memory_graph}. Each session is passed to the extractor $g_{\text{EDU}}$ along with timestamps and speaker tags, and the model emits extracted EDUs with source turn index attributions. 
No dataset-specific modification is required.

\paragraph{Assistant utterances: atomic EDUs and structured chunks.}
To mitigate the above discussed issues, each assistant turn is processed in two parallel views:

\begin{compactenum}
    \item \textbf{Atomic EDUs.} The extractor produces a set of fine-grained EDUs, analogous to the user side, each with its own source turn index. These capture localized facts (e.g., an atomic fact or an event).
    \item \textbf{Structured chunks.} In the same call, the extractor is asked to identify cohesive information blocks---``structured chunks''---that group related details presented in an organized way (comparisons, detailed overviews, comprehensive recommendations, step-by-step procedures, lists of related items, etc.). For each chunk $c$, the model also generates a short summary $s(c)$ of 2--3 sentences that (i) states which user request or question the chunk addresses, (ii) describes the main information categories covered, and (iii) naturally includes key entities and terms.
\end{compactenum}

We treat each chunk as an additional EDU node in the memory graph, but use the summary $s(c)$ as the textual content $\text{text}(e)$ for argument extraction, indexing, and retrieval. 
Arguments for chunk nodes are extracted from $s(c)$ rather than from the full chunk text. 
The original chunk content is stored separately and is only revealed to the QA model at answer time when the corresponding EDU node is retrieved. 
This design preserves the organizational structure of long assistant responses while keeping the indexable text short and information-dense.

In practice, both atomic EDUs and structured chunks share the same metadata schema, including source turn indices and session timestamps, and are handled uniformly by the retrieval and graph components. 
The only difference is that chunk nodes have a hidden ``expanded'' text used solely in the final QA prompt and we need to initiate two LLM calls on this dataset. 
This dataset-specific adaptation improves recall on LongMemEval$_\text{S}$ without changing the core \Ours/\OursGraph algorithms.

\end{document}